\documentclass[10pt,conference]{IEEEtran}
\IEEEoverridecommandlockouts

\usepackage{bbding}
\usepackage{caption}

\usepackage{graphicx}
\usepackage[many]{tcolorbox}

\usepackage{mathtools,amssymb,latexsym,amsfonts,stmaryrd,bm}
\usepackage{pbox}
\usepackage{xfrac}
\usepackage{booktabs}
\usepackage{xcolor}
\usepackage{threeparttable}
\usepackage{amsthm}
\usepackage{bbding}
\usepackage{multirow}
\usepackage{tikz}
\usepackage{mathpartir}
\usepackage[ruled,linesnumbered]{algorithm2e}
\usepackage{url}
\usepackage{syntax}
\usepackage{framed}
\usepackage[noend]{algpseudocode}
\usepackage{flushend}
\usepackage{listings}
\usepackage{moresize}
\usepackage{wrapfig}
 
\usepackage{seqsplit}
\usepackage{alltt}
\usepackage{xspace}
\usepackage{enumitem}
\usepackage{pifont}
\usepackage{makecell}

\DeclareMathAlphabet{\mathcal}{OMS}{cmsy}{m}{n}

\usepackage{amsmath, listings, amsthm, amssymb, proof, xspace}

\usepackage{thmtools}

\declaretheoremstyle[spaceabove=\topsep,notefont=\normalfont\itshape]{mystyle}

\newcommand{\revise}[2]{{\color{red}{\ifx&#1&\else- #1\fi}} {\color{ForestGreen}{\ifx&#2&\else+ #2\fi}}}%
\renewcommand{\revise}[2]{#2}%

\usetikzlibrary{arrows,patterns, decorations.pathreplacing}
\newtheorem{definition}{Definition}
\newcommand{\Rom}[1]{(\uppercase\expandafter{\romannumeral #1\relax})}

\newcommand{\F}{Fig.}

\newcommand{\T}{Table}
\renewcommand{\S}{Sec.}
\newcommand{\A}{Alg.}

\newcommand{\ignore}[1]{}

\lstdefinestyle{base}{
  moredelim=**[is][\color{red}]{@}{@},
  escapeinside={<@}{@>}
}

\newcommand{\parh}[1]{\noindent\textbf{#1}}
\newcommand{\sparh}[1]{\noindent\textbf{#1}}

\usepackage{amssymb}
\usepackage{pifont}

\newcommand\DejaVuttfamily{%
  \fontfamily{DejaVuSansMono-TLF}\selectfont }

\lstdefinestyle{base}{
  moredelim=**[is][\color{red}]{@}{@},
  escapeinside={<@}{@>}
}

\lstdefinelanguage
   [x64]{Assembler}     
   [x86masm]{Assembler} 
   {morekeywords={CDQE,CQO,CMPSQ,CMPXCHG16B,JRCXZ,LODSQ,MOVSXD, %
                  POPFQ,PUSHFQ,SCASQ,STOSQ,IRETQ,RDTSCP,SWAPGS, %
                  rax,rdx,rcx,rbx,rsi,rdi,rsp,rbp, %
                  r8,r8d,r8w,r8b,r9,r9d,r9w,r9b}} 

\usepackage{listings}
\usepackage{fancyvrb}
\usepackage{color}
\definecolor{lightgray}{rgb}{.9,.9,.9}
\definecolor{darkgray}{rgb}{.4,.4,.4}
\definecolor{purple}{rgb}{0.65, 0.12, 0.82}
\definecolor{commentgreen}{RGB}{63,127,95}
\definecolor{redMark}{RGB}{252, 228, 214}
\definecolor{yellowMark}{RGB}{255, 242, 204}
\definecolor{greenMark}{RGB}{226, 239, 218}
\definecolor{blueMark}{RGB}{221, 235, 247}
\definecolor{lightred}{RGB}{255, 130, 130}
\definecolor{myGreen}{RGB}{61, 161, 61}
\definecolor{myBrown}{RGB}{196, 89, 17}

\colorlet{myPurple}{blue!40!red}
\definecolor{myOrange}{RGB}{255,192,0}
\newcommand{\code}[1]{\textcolor{blue}{\textit{\bfseries{#1}}}}

\newcommand{\enc}[1]{$\phi^{*}_{\theta}$}
\newcommand{\dec}[1]{$\psi^{*}_{\theta}$}

\lstdefinelanguage{Solidity}{
  keywords={len,delete,int,void,payable, public, event, contract, typeof, new, true, false, catch, function, return, null, catch, switch, var, if, in, while, do, else, case, break,struct,const,socklen_t,sa_familty_t,char,sockaddr},
  keywordstyle=\color{violet}\bfseries,
  ndkeywords={class, export, boolean, throw, implements, import, this},
  ndkeywordstyle=\color{darkgray}\bfseries,
  identifierstyle=\color{black},
  sensitive=false,
  comment=[l]{//},
  escapeinside={(*@}{@*)},          
  morecomment=[s]{/*}{*/},
  commentstyle=\color{commentgreen}\ttfamily,
  stringstyle=\color{red}\ttfamily,
  morestring=[b]',
  morestring=[b]"
}

\newcommand{\rnum}[1]{\uppercase\expandafter{\romannumeral #1\relax}}

\usepackage{xcolor}

\algnewcommand{\LeftComment}[1]{\Statex \(\triangleright\) #1}

\definecolor{pptbrown}{RGB}{132,60,12}
\definecolor{pptgreen}{RGB}{56,87,35}

\let\OLDthebibliography\thebibliography
\renewcommand\thebibliography[1]{
  \OLDthebibliography{#1}
  \setlength{\parskip}{0pt}
  \setlength{\itemsep}{0pt plus 0.1ex}
}

\definecolor{pptgreen}{RGB}{84,130,53}
\definecolor{pptred}{RGB}{176,35,24}
\definecolor{pptblue}{RGB}{194,214,236}

\definecolor{pptgreen1}{RGB}{78,173,91}
\definecolor{pptred1}{RGB}{192,0,0}

\definecolor{pptyellow1}{RGB}{203,195,167}
\definecolor{pptgreen2}{RGB}{184,192,176}

\xspace

\usepackage[colorlinks]{hyperref}   
\AtBeginDocument{%
  \hypersetup{pdfborder={0 0 1}, urlcolor=.}  
}

\definecolor[named]{ACMBlue}{cmyk}{1,0.1,0,0.1}
\definecolor[named]{ACMYellow}{cmyk}{0,0.16,1,0}
\definecolor[named]{ACMOrange}{cmyk}{0,0.42,1,0.01}
\definecolor[named]{ACMRed}{cmyk}{0,0.90,0.86,0}
\definecolor[named]{ACMLightBlue}{cmyk}{0.49,0.01,0,0}
\definecolor[named]{ACMGreen}{cmyk}{0.20,0,1,0.19}
\definecolor[named]{ACMPurple}{cmyk}{0.55,1,0,0.15}
\definecolor[named]{ACMDarkBlue}{cmyk}{1,0.58,0,0.21}
\hypersetup{
  colorlinks,
  linkcolor=ACMPurple,
  citecolor=ACMPurple,
  urlcolor=ACMDarkBlue,
  filecolor=ACMDarkBlue
}

\usepackage{inconsolata} 
\usepackage{biolinum} 

\usepackage{microtype}

\usepackage[noadjust]{cite}

\usepackage{soul}
\usepackage[normalem]{ulem}

\newif\ifshowcomments
\showcommentstrue

\ifshowcomments
\newcommand{\pc}[1]{\mytodored{[Pingchuan: #1]}}
\else
\newcommand{\pc}[1]{}
\fi
\newcommand{\mytodored}[1]{\textcolor{red}{\ding{46}~{\sf}~#1}}

\def\BibTeX{{\rm B\kern-.05em{\sc i\kern-.025em b}\kern-.08em
    T\kern-.1667em\lower.7ex\hbox{E}\kern-.125emX}}

\begin{document}

\title{Causality-Aided Trade-off Analysis for Machine Learning Fairness}

\author{
    \IEEEauthorblockN{
        Zhenlan Ji\IEEEauthorrefmark{1},
        Pingchuan Ma\IEEEauthorrefmark{1}\IEEEauthorrefmark{3},
        Shuai Wang\IEEEauthorrefmark{1}\IEEEauthorrefmark{3},
        and Yanhui Li\IEEEauthorrefmark{2}
    }\thanks{\IEEEauthorrefmark{3} Corresponding authors}
    \IEEEauthorblockA{
        \IEEEauthorrefmark{1}The Hong Kong University of Science and Technology,
        \{zjiae, pmaab, shuaiw\}@cse.ust.hk
    }
    \IEEEauthorblockA{
        \IEEEauthorrefmark{2}State Key Laboratory for Novel Software Technology, Nanjing University,
        yanhuili@nju.edu.cn
    }
}

\maketitle

\begin{abstract}

There has been an increasing interest in enhancing the fairness of machine
learning (ML). Despite the growing number of 
fairness-improving methods, we lack a systematic understanding of the
trade-offs among factors considered in the ML pipeline when fairness-improving
methods are applied. This understanding is essential for developers to make
informed decisions regarding the provision of fair ML services. Nonetheless, it
is extremely difficult to analyze the trade-offs when there are multiple
fairness parameters and other crucial metrics involved, coupled, and even in
conflict with one another.

This paper uses causality analysis as a principled method for analyzing
trade-offs between fairness parameters and other crucial metrics in ML
pipelines. To practically and effectively conduct causality analysis, we propose
a set of domain-specific optimizations to facilitate accurate causal
discovery and a unified, novel interface for trade-off analysis based on
well-established causal inference methods. We conduct a comprehensive empirical
study using three real-world datasets on a collection of widely-used
fairness-improving techniques. Our study obtains actionable suggestions for
users and developers of fair ML. We further demonstrate the versatile usage of
our approach in selecting the optimal fairness-improving method, paving the way
for more ethical and socially responsible AI technologies.

\end{abstract}

\section{Introduction}
\label{sec:introduction}


Machine learning (ML) techniques are now essential for everyday applications in
safety-critical domains like credit risk evaluation~\cite{bono2021algorithmic}
and criminal justice~\cite{berk2021fairness}. However, ML models have exhibited
inherent biases~\cite{pleiss2017fairness,peng2022fairmask}, leading to
real-world consequences such as discriminatory outcomes between privileged and
underprivileged groups~\cite{zhang2018mitigating, agarwal2018reductions,
chakraborty2020fairway}. To address this, various \textit{fairness-improving
methods} have been proposed and studied by the software engineering (SE)
community, including
mitigating unfairness through data
processing~\cite{chakraborty2021bias,li2022training,peng2022fairmask}, model
modification~\cite{gao2022fairneuron, tao2022ruler}, or prediction
alteration~\cite{pleiss2017fairness}.

Despite the significant progress made, an important question arises:
\textit{what are the trade-offs made by these fairness-improving methods in the
ML pipeline?} It is a widely held belief that there exists a trade-off between
fairness and the functional quality properties of the ML pipeline, such as the
ML performance and the ML model robustness. In general, empirical studies from
the SE community and theoretical analyses from the ML community have
demonstrated that optimizing for performance may come at the cost of fairness,
and vice versa~\cite{biswas2020machine, chakraborty2021bias, li2022training,
friedler2016possibility, binns2020apparent}. Furthermore, many metrics
concentrating on fairness, such as group fairness and individual fairness, are
inherently incompatible~\cite{friedler2016possibility}. These trade-offs render
additional complexity to the process of improving fairness in ML systems and are
not well understood. As a result, the lack of transparent and manageable
trade-off analyses makes it challenging for developers to make informed
decisions within the ML pipeline.


To understand trade-offs, it is essential to comprehend the interactions among
fairness-improving methods as well as different metrics. More importantly, it is
crucial to ``disentangle'' the true cause-effect relationships from the observed
correlations. For example, a fairness-improving method may simultaneously affect
the model's fairness on both the training set and the test set. However, it is
unclear how would training fairness affect test fairness due to the confounding
influence introduced by the fairness-improving method. To hurdle this obstacle,
we advocate for the use of causality analysis~\cite{pearl2009causality} --- a
principled approach to learning the causal relations between random variables
--- to better understand trade-offs between fairness parameters and other
crucial metrics in the ML pipeline when fairness-improving methods are
enforced.

Despite the promising potential of causality analysis, several challenges must
be addressed to fully harness its power. A typical causality analysis procedure
can be broadly divided into two phases: \ding{192} causal graph learning, and
\ding{193} causal inference. In the first phase (\ding{192}), a causal graph is
learned from data, whose nodes essentially represent random variables (which are
fairness parameters and other metrics in the ML pipeline) and the edges encode
the causal relationships among nodes. The second phase (\ding{193}) involves
applying an inference algorithm to the learned causal graph to quantitatively
estimate the causal effect from one node to another. However, both phases
present considerable challenges in our research context. First, when treating
fairness parameters and other metrics in the ML pipeline as variables, the
process of causal graph learning becomes a causal discovery problem involving
mixed data types, such as continuous and discrete variables. Furthermore, the
complex nature of ML pipelines, which involve numerous metrics and methods,
makes learning an accurate causal graph challenging due to the well-known
``curse of dimensionality'' problem. Existing causal discovery algorithms have
difficulty managing cases with this level of complexity. Second, even if a
causal graph can be successfully learned, leveraging the graph to understand
trade-offs between fairness-improving techniques and metrics remains
under-explored. Specifically, it is unclear how to recast trade-off analyses
into a series of standard causal inference queries such that well-established
causal inference algorithms can be applied smoothly.

To address these challenges, we propose a novel causal analysis framework for
understanding trade-offs. In the causal graph learning phase (\ding{192}), we
involves fairness-improving methods as additional interventional variables to
guide the learning process, and introduce a novel mechanism to convert discrete
variables into continuous one without losing information. In the causal
inference phase (\ding{193}), we systematically formulate trade-offs in typical
ML pipelines using causality analysis. Our formulation provides a novel and
unified interface over various fairness-improving methods and critical metrics
in the ML pipeline, covering both the training and testing phases. 

To gauge the effectiveness of our approach, we conduct extensive experiments
using three real-world datasets: Adult~\cite{adult}, COMPAS~\cite{compas}, and
German~\cite{german} with 12 widely-used fairness-improving methods used (see
\T~\ref{tab:fair-methods}). This empirical study enables a comprehensive
analysis of the trade-offs and leads to a number of intriguing findings. First,
the selection of fairness metrics can significantly affect the pattern of
observed trade-offs, highlighting the need for a systematic and automated
approach to deciding optimal metrics for a particular scenario. (b) Second,
certain metrics, such as Average Odds Difference (AOD) and Theil Index (TI),
play a central role in trade-offs (including fairness vs. performance and
fairness vs. robustness). These metrics act as the cause for trade-offs more
frequently than other metrics. Third, the trade-off between fairness and
robustness, though not extensively explored in the SE community, is inevitable.
This observation highlights the importance of taking both robustness and
fairness into account when calibrating the ML pipeline. Furthermore, we
demonstrate a versatile application of our framework in the selection of optimal
fairness-improving methods. Empirical results in \S~\ref{sec:application}
indicate that our approach outperforms state-of-the-art methods.
To conclude, this paper makes the following contributions: 



\begin{itemize}[leftmargin=*,topsep=0pt,itemsep=0pt]
	\item To our knowledge, this is \textit{the first work} to introduce
	causality analysis as a principled approach to analyzing trade-offs between
	fairness and other critical metrics in ML pipelines.
	\item We propose a novel causality analysis framework to practically and
	effectively concretize causality analysis in the context of ML pipelines
	when fairness-improving methods are enforced. In particular, we deliver a
	set of domain-specific optimizations to enable more accurate causal
	discovery and design a unified interface for trade-off analysis on the basis
	of standard causal inference techniques.
	\item We conduct an extensive empirical study on representative
	fairness-improving methods and real-world datasets. We obtain actionable
	suggestions for users and developers in the ML pipeline when
	fairness-improving methods are enforced.
\end{itemize}

\parh{Open Source.} The source code and data are available at~\cite{artifact}.




\section{Preliminary}
\label{sec:background}

\begin{figure*}[t]
    \centering
    \includegraphics[width=0.9\linewidth]{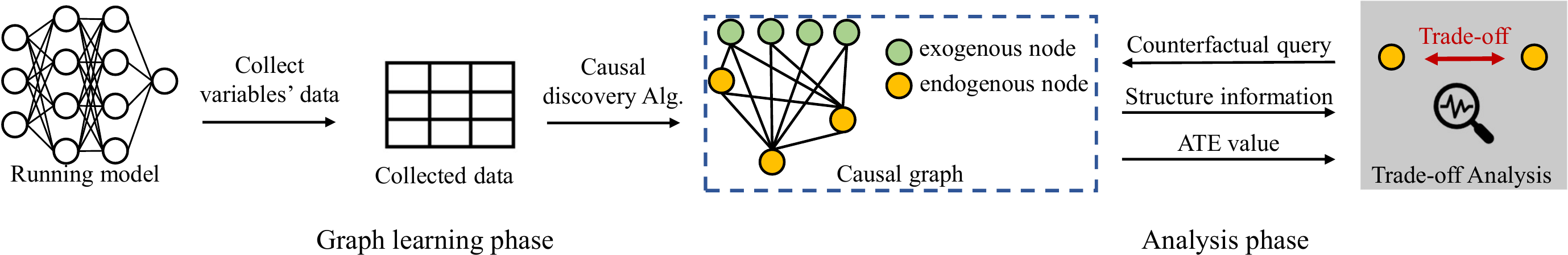}
    \vspace{-10pt}
    \caption{Study overview.}
    \label{fig:workflow}
    \vspace{-10pt}
\end{figure*}

\subsection{Fairness-Improving Methods}
\label{subsec:fairness}

A significant amount of research has been devoted to the investigation of
fairness in ML~\cite{kamiran2012data, kamishima2012fairness,
kamiran2012decision, feldman2015certifying, rozsa2016accuracy,
pleiss2017fairness, zhang2018mitigating, agarwal2018reductions,
chakraborty2020fairway, ma2020metamorphic, chakraborty2021bias,
peng2022fairmask, li2022training, chen2022maat, MonjeziTTT23, ma2023oops,
majumder2021fair, chen2023comprehensive}. In general, most
fairness-improving methods can be classified according to the ML stages
when they are involved and the fairness objectives they aim to
achieve~\cite{biswas2020machine, zhang2022adaptive}.

\parh{Methods by Different Stages.}~Fairness-improving methods can be
categorized based on the stage of the machine learning pipeline they are
involved in. These stages can be classified as pre-processing, in-processing, or
post-processing methods. Pre-processing methods manipulate the training data,
while in-processing methods modify the model during training. Post-processing
methods, on the other hand, adjust the predictions made by the trained model.
Representative methods for each type of fairness-improving method have been
selected and analyzed in our study (see \T~\ref{tab:fair-methods} for more
information).

\parh{Methods by Different Objectives.}~Fairness-improving methods can also
be categorized based on the fairness objectives they aim to achieve.
Generally, there are two primary types of fairness objectives: individual
fairness and group fairness. Individual fairness refers to treating all
individuals equally. Zhang et al.~\cite{zhang2022adaptive} introduced the
Causal Discrimination Score (CDS) to quantify the individual fairness of a
model. On the other hand, group fairness focuses on the fairness of a
model's predictions for different groups defined by one or more sensitive
attributes. A model is considered fair if it treats all subpopulations
equally. Common group fairness metrics include Disparate Impact
(DI)~\cite{feldman2015certifying} and Statistical Parity Difference
(SPD)~\cite{calders2010three}, as presented in Table~\ref{tab:metrics}.

\parh{Motivation.}~The different types of fairness-improving methods, along with
their varied objectives, can introduce complexity and challenges in achieving a
balanced solution. For instance, two well-known fairness objectives, group
fairness and individual fairness, are often mutually incompatible, leading to an
inevitable trade-off between them~\cite{friedler2016possibility,
binns2020apparent}. A similar phenomenon is also observed in other objectives
such as robustness versus accuracy~\cite{tsipras2018robustness}. \textit{These
complex relationships behind potentially conflicting objectives render achieving
a good trade-off among multiple metrics (fairness, accuracy, robustness)
challenging.}
Overall, this research advocates the usage of causality analysis, 
a well-established and systematic approach, for understanding the complex
relationships between variables presented in this fairness-improving context
over ML pipeline. We envision that, by employing causality analysis, we can make
the process of improving fairness notably more transparent and manageable. We
present the preliminary knowledge of causality analysis in the following
subsection.

\subsection{Causality Analysis}
\label{subsec:causality}

Causality analysis effectively provides a systematic and comprehensive
understanding of the complex causal relationships between
variables~\cite{dubslaff2022causality,baluta2022membership, ma2022xinsight,
ji2022perfce}. This desirable feature is attained through two fundamental
steps: causal discovery and causal inference.


According to Judea Pearl~\cite{pearl2009causality}, causation (or causal
relation) refers to the relationship between two variables, wherein changes
in one variable cause changes in the other. This concept differs from
correlation, which merely indicates the statistical dependence between two
variables. Considering a simple example, $X \leftarrow Z\rightarrow Y$,
where $X$ and $Y$ are correlated, but $X$ does not cause $Y$ (and vice
versa). Here, $Z$ is a \textit{confounder} because it simultaneously causes
$X$ and $Y$. Indeed, the correlation between $X$ and $Y$ is induced by $Z$.
This example highlights why correlation alone cannot imply causation. To
address more complex cases, we introduce the definition of \textit{causal
graph} as follows:


\vspace{-5pt}
\begin{definition}[Causal Graph]
    A causal graph (a.k.a., Bayesian network) is a directed acyclic graph
    (DAG) consisting of nodes $V$ and edges $E$, i.e., $G = (V, E)$, where
    each node $X$ ($X \in V$) represents a random variable and each edge $X
    \rightarrow Y$ ($X, Y \in V$) represents a directed causal relation
    from $X$ to $Y$. 
    The nodes in the graph can be categorized into two groups:
    \textit{endogenous} nodes, which are determined by the values of other
    nodes in the graph, and \textit{exogenous} nodes, which are determined
    by external factors.
\end{definition}
\vspace{-5pt}

Causal graphs facilitate reasoning both qualitative and quantitative causal
relations between variables. In practice, however, the causal graph is commonly
\textit{unknown}. For another, although causal relationships can be inferred if
interventions are properly applied, the majority of real-world variables cannot
be simply intervened~\cite{glymour2019review}. This obstacle necessitates causal
discovery, which aims to reconstruct the causal graph from observational data.

\parh{Causal Discovery.}~Causal discovery is the process of inferring a directed
acyclic graph (DAG), where each node represents a variable and each edge
represents a causal relation. Holistically, mainstream causal discovery methods
can be categorized into three groups: constraint-based, score-based, and
model-based methods~\cite{glymour2019review}. Constraint-based methods use the
\textit{conditional independence test} to determine the edges and properties of
special relationships (such as \textit{confounder}) to infer the direction of
causal relations~\cite{spirtes2000constructing, spirtes2000causation,
colombo2012learning, ma2022ml4s, wang2023towards, ding2020reliable}. Score-based
methods formulate causal discovery as a search problem and evaluate the quality
of the causal graph using a score function~\cite{chickering2002optimal,
ogarrio2016hybrid, lorch2021dibs, ma2022noleaks}. For model-based methods,
asymmetry is exploited to identify causal relations~\cite{shimizu2006linear,
shimizu2011directlingam, hoyer2008nonlinear}.



\parh{Causal Inference.}~Causal inference quantitatively estimates the
causal effect of one variable on another variable based on a causal graph.
Although the causal graph is generally interpretable, estimating the causal
effect can be challenging. Suppose, for instance, there are three variables
$X$, $Y$, and $Z$, where $Z \rightarrow X \rightarrow Y$ and meanwhile $Z
\rightarrow Y$. Here, it is challenging to accurately determine the true
causal effect of $X$ on $Y$ while minimizing the impact of $Z$. In
causality analysis, Average Treatment Effect (ATE) is commonly utilized to
address this issue~\cite{pearl2009causality}. Below is the definition of
ATE:

\vspace{-5pt}
\begin{definition}[ATE]
    In causal graph $G$, ATE of $X$ on $Y$ can be computed as:
    \begin{equation}
        \textnormal{ATE} = \mathbb{E}[Y \mid do(X=\bm{x}_1)] - \mathbb{E}[Y \mid do(X=\bm{x}_2)]
        \label{eq:ate}
    \end{equation}
    where the $do(\cdot)$ operator denotes a counterfactual query, which
    represents a hypothetical intervention over the value of a variable
    $X$ (i.e., $X$ is set to a constant value $\bm{x}$, which may not be 
    observed in the data). $\bm{x}_1$ and $\bm{x}_2$ are two arbitrary
    values of $X$ that are determined by the user.
\end{definition}
\vspace{-5pt}

Since ATE employs counterfactual queries, it cannot be explicitly estimated
from observational data as it is a \textit{causal estimand}. The process of
making an incomputable causal estimand computable is known as
\textit{causal inference}~\cite{neal2015introduction}.
This paper uses a popular causal estimation method, double machine learning
(DML)~\cite{chernozhukov2016double}. Continuing with the preceding example,
where $Z \rightarrow X \rightarrow Y$ and meanwhile $Z \rightarrow Y$. $X$
denotes the treatment variable, $Y$ denotes the outcome variable, and $Z$
denotes the confounder.\footnote{For simplicity, we assume that they all
represent a single variable, although, in more general scenarios, they may
also be regarded as a set of variables.} DML uses two arbitrary machine
learning models to estimate $X$ and $Y$ from $Z$, respectively. The effect
of the confounder $Z$ can then be ``removed'' by estimating the difference
between the predicted and observed values of $X$ and $Y$ (i.e., the
residuals). DML makes no assumptions about the form of the confounder $Z$'s
effect~\cite{econml}, making it applicable to a wide range of causal
relations.




\section{Study Pipeline}
\label{sec:design}

\F~\ref{fig:workflow} depicts our study workflow, which consists of two main
phases: (1) graph model construction and (2) trade-off analysis.
In the first phase, we collect a substantial quantity of data, which includes
un-interventional observed metrics such as test accuracies and SPD scores, and
interventional metrics like the fairness-improving methods' parameters. 
Certain variables are referred to as ``un-interventional'' because their
values cannot be directly changed by the user (e.g., the model test accuracy).
We then use a causal discovery algorithm to learn a causal graph, where nodes
represent variables (metrics and parameters)\footnote{In our context, exogenous
nodes only represent the user-determined parameters of fairness-improving
methods or model training, and endogenous nodes represent the observed metrics.
Hence, with a slight abuse of notation, exogenous nodes are also referred to as
interventional nodes, and endogenous nodes are referred to as observational
(un-interventional) nodes in this paper.}, and directed edges represent causal
relations between them. 

In the second phase, we propose counterfactual queries according to the
identified trade-off between two un-interventional nodes on the causal graph;
these nodes typically include the metrics of model accuracy, fairness, and
robustness. Our aim is to explain the underlying cause for the trade-off among
these important metrics, thus providing insights into the influence of the ML
fairness-improving methods over other important factors on the ML pipeline.

\subsection{Graph Model Construction}
\label{subsec:graph-build}

This study seeks to reveal the intricate relationships among diverse kinds of
metrics. However, the sheer number of metrics and the complexity of the
relationships between them pose considerable obstacles in learning a
sufficiently precise causal graph. \ding{202} and \ding{203} illustrate how we
address this challenge from the perspectives of data collection and graph
learning.

\sparh{\ding{202} Collecting Training Data.}~Due to the limited number of
variables (e.g., only 14 nodes are involved in Baluta et al.'s
work~\cite{baluta2022membership}), prior works tend to focus on
un-interventional variables. In contrast, because of the large number of
variables involved in this problem, \cite{baluta2022membership}'s practice fails
to guarantee that the entire value range of each variable is exhaustively
covered. Therefore, we introduce fairness-improving methods as interventional
nodes in the causal graph. For parameter-tunable methods, we convert and
normalize their parameters into a ratio, which ranges from 0 to 1. For
non-parameter-tunable methods, we use probabilistic sampling, also converting
them into ratios as the following equation shows.
\begin{equation}
    D' = T_{fair}(\alpha D) \cup (D-\alpha D)
    \label{eq:probabilistic}
\end{equation}
where $D$ is the original dataset, $T_{fair}$ is the fairness-improving
method, $\alpha$ denotes the ratio, $\alpha D$ presents the selected
$\alpha$ part of the original dataset, $(D-\alpha D)$ denotes the remaining
part, and $D'$ is the resulting dataset by applying $T_{fair}$ on the
$\alpha$ part. Accordingly, all fairness-improving methods can be
represented as a ratio and can be treated as nodes in the causal graph. By
intervening these nodes, we can collect sufficient, high-quality training
data for causal discovery, with all possible value ranges of
un-interventional variables covered.


\sparh{\ding{203} Learning Causal Graph.}~We employ DiBS~\cite{lorch2021dibs}, a
state-of-the-art score-based method, to learn causal graph with variational
inference and is more efficient than other methods that rely on Markov chain
Monte Carlo (MCMC) sampling. As will be shown in \S~\ref{sec:pilot}, DiBS is
able to accurately learn causal graphs and aligns well with expert knowledge.

\subsection{Trade-off Analysis}

From a holistic perspective, the influence of a specific fairness-improving
method can be considered as a change propagation throughout the causal graph.
Initially, the parameters of fairness-improving methods, denoted as a node on
the causal graph, directly affect their children nodes, which subsequently leads
to indirect impacts on their descendants. As a consequence of this influence
propagation, various descendants might experience either improvements or
downgrades.

When examining a pair of metrics ($X$ and $Y$) that are affected
distinctively by the fairness-improving method, it becomes essential to
comprehend the trade-off between $X$ and $Y$. Although the causal graph has
offered a qualitative view of the relationships between metrics and
fairness-improving methods (e.g., one method affects a metric), it is not
sufficient to accurately quantify the trade-off involving two specific
metrics. For instance, we may first observe ``Test Accuracy'' decreases
while ``Test SPD'' (Statistical Parity Difference) increases after
applying, Reweighing~\cite{kamiran2012data}, a fairness-improving method
(see details in \T~\ref{tab:fair-methods}). Then, on the causal graph, we
observe that ``Test Accuracy'' and ``Test SPD'' has a common ancestor
``Dataset SPD'' (SPD of the dataset). However, without further analysis, we
are unable to conclude if the trade-off is caused by ``Dataset SPD'' or by
other factors (e.g., other common ancestors). In the following section, we
will first provide a formal definition of the trade-off that occurs between
two metrics. Then, we will show how a trade-off can be explained by the
combination of causal graphs and ATE analysis.


\parh{Trade-off Definition.}~In line with prior works~\cite{speicher2018unified,
hort2021fairea, zhang2022adaptive}, we define the concept of \textit{trade-off}:
a situation where an effort to \textit{enhance} one aspect of a system (i.e., an
endogenous node in our research context) results in a \textit{downgrade} in
another aspect. The quantifiable effect of one metric on another can be
precisely measured using average treatment effect (ATE), enabling the
identification of potential causes for the trade-off.

To adequately define whether a change in one metric is an enhancement or
downgrade, we introduce a user-defined function
$\code{sign}:\mathbb{R}\times\mathbb{R}\to \{+,-\}$. This function takes the
current value and the change of a metric as input, and outputs the corresponding
sign of the change (``$+$'' indicating an improvement and ``$-$'' indicating a
downgrade). It is worth noting that the definition of $\code{sign}(\cdot,\cdot)$
is specific to each metric. For instance, in the case of accuracy,
$\code{sign}_{\text{Acc}}(0.6,0.1)$ yields ``$+$'' to demonstrate that the
accuracy has improved due to the change. Conversely, for the SPD score,
$\code{sign}_{\text{SPD}}(0.6,0.1)$ yields ``$-$'' to illustrate that the
statistical parity difference has been downgraded as a result of the change.
Then, given two metrics $X$ and $Y$ and a fairness-improving method $T$, there
is a trade-off between $X$ and $Y$ if and only if $\code{sign}_X(x_{T=0},
\text{ATE}^{x}_{t}) \neq \code{sign}_Y(y_{T=0},\text{ATE}^{y}_{t})$, where
$x_{T=0}$ and $y_{T=0}$ are the values of $X$ and $Y$ before applying $T$,
respectively, and $\text{ATE}^{x}_{t}$ and $\text{ATE}^{y}_{t}$ are the ATEs of
$T$ on $X$ and $Y$, respectively. 
This condition indicates that the fairness-improving method $T$ has distinct
effects on $X$ and $Y$. 


\begin{algorithm}[!htbp]
    \scriptsize
    \caption{Analysis}
    \label{alg:analysis}
    \KwIn{Causal Graph $G$, Fairness Improving Method $T$, two Metrics $X$ and $Y$.}
    \KwOut{Trade-off causes list $C$}
    $C \leftarrow \emptyset$\;
    $\text{ATE}^{x}_{t} \leftarrow \mathbb{E}[X \mid do(T=1)] - \mathbb{E}[X \mid do(T=0)]$\;
    $\text{ATE}^{y}_{t} \leftarrow \mathbb{E}[Y \mid do(T=1)] - \mathbb{E}[Y \mid do(T=0)]$\;
    $x_{T=0} \leftarrow \mathbb{E}[X|T=0]$; $y_{T=0} \leftarrow \mathbb{E}[Y|T=0]$\;
    \If{\normalfont $\code{sign}_X(x_{T=0}, \text{ATE}^{x}_{t}) = \code{sign}_Y(y_{T=0},\text{ATE}^{y}_{x}$)}{
        \Return{$\bot$}; // Terminate the algorithm in the absence of trade-offs.\\
    }
    \If {$X$ causes $Y$ on $G$}{
        $x_{T=1} \leftarrow \mathbb{E}[X|T=1]$\; 
        $\text{ATE}^{y}_{x} \leftarrow \mathbb{E}[Y \mid do(X={x_{T=1}})] - \mathbb{E}[Y \mid do(X=x_{T=0})]$\;
        \lIf {\normalfont $\code{sign}_X(x_{T=0},\text{ATE}^{x}_{t}) \neq \code{sign}_Y(y_{T=0},\text{ATE}^{y}_{x})$}{
            $C \leftarrow C \cup \{X\}$
        }
    }
    \ElseIf {$Y$ causes $X$ on $G$}{
        $y_{T=1} \leftarrow \mathbb{E}[Y|T=1]$\;
        $\text{ATE}^{x}_{y} \leftarrow \mathbb{E}[X \mid do(Y=y_{T=1})] - \mathbb{E}[X \mid do(Y=y_{T=0})]$\;
        \lIf {\normalfont $\code{sign}_Y(y_{T=0},\text{ATE}^{y}_{t}) \neq \code{sign}_X(x_{T=0},\text{ATE}^{x}_{y})$}{
            $C \leftarrow C \cup \{Y\}$
        }
    }
    $P_{C} \leftarrow \text{FindPotentialCause}(G, X, Y)$\;
    \ForEach{$P \in P_{C}$}{
        $p_{T=1} \leftarrow \mathbb{E}[P|T=1]$; $p_{T=0} \leftarrow \mathbb{E}[P|T=0]$\;
        $\text{ATE}^{x}_{p} \leftarrow \mathbb{E}[X \mid do(P=p_{T=1})] - \mathbb{E}[X \mid do(P=p_{T=0})]$\;
        $\text{ATE}^{y}_{p} \leftarrow \mathbb{E}[Y \mid do(P=p_{T=1})] - \mathbb{E}[Y \mid do(P=p_{T=0})]$\;
        \lIf {\normalfont $\code{sign}_X(x_{T=0},\text{ATE}^{x}_{p}) \neq \code{sign}_Y(y_{T=0},\text{ATE}^{y}_{p})$}{
            $C \leftarrow C \cup \{P\}$
        }
    }
    \Return{$C$}
\end{algorithm}

\parh{Trade-off Analysis.}~The process of performing trade-off analysis is
described in \A~\ref{alg:analysis}. 
Overall, \A~\ref{alg:analysis} accepts a causal graph $G$ (generated in
\ding{203} of \S~\ref{subsec:graph-build}), a fairness-improving method
$T$, and two metrics $X$ and $Y$ (lines 1--2). It returns a list of
identified causes for the trade-off between $X$ and $Y$ (line 25), where
each cause (can be $X$ or $Y$ itself) represents one node on the causal
graph.
Following the above trade-off definition and the usage of $T=0$, we note that
$T=1$ in \A~\ref{alg:analysis} indicates that the fairness-improving method $T$
is applied. 

When provided with two metrics $X$ and $Y$, \A~\ref{alg:analysis} initially
checks for the existence of a trade-off between them, based on the trade-off
definition we just present (lines 2--7). And if a trade-off is present, the
algorithm proceeds. 
Subsequently, if $X$ is a cause of $Y$ (i.e., there exists a path from $X$ to
$Y$ in the causal graph $G$), \A~\ref{alg:analysis} computes the ATE of $X$ on
$Y$ (lines 9--10) and verifies whether the trade-off is caused by $X$ (lines
11--12). Likewise, if $Y$ is a cause of $X$, \A~\ref{alg:analysis} repeats
the process with reversed roles (lines 13--17).
%
%
Finally, \A~\ref{alg:analysis} queries the causal graph to identify the
potential causes of the trade-off, which are the common ancestors of $X$ and $Y$
(line 18). For each potential cause (line 19), \A~\ref{alg:analysis} calculates
its ATE on both $X$ and $Y$ (lines 21--22). If the ATEs implies a trade-off, the
potential cause is designated as a cause (line 23).




\section{Experiment Setup}
\label{sec:setup}

Our study is implemented in Python with roughly 2.8K lines of code. All
experiments are launched on one AMD CPU Ryzen Threadripper 3970X and one
NVIDIA GPU GeForce RTX 3090.

\begin{table}[!htpb]
    \centering
    \scriptsize
    \caption{Dataset information.}
    \label{tab:data}
    \setlength{\tabcolsep}{3.0pt}
    \resizebox{1.0\linewidth}{!}{
        \begin{tabular}{l|c|c|c|c}
            \hline
            \multirow{2}{*}{\textbf{Dataset}}                      & \multirow{2}{*}{\textbf{Size}}                      & \textbf{Favorable}       & \textbf{Sensitive} & \textbf{Privileged} \\
            && \textbf{Class}& \textbf{Attribute} &  \textbf{Group} \\\hline
            \multirow{2}{*}{Adult~\cite{adult}}   & \multirow{2}{*}{48,842$\times$12 } & \multirow{2}{*}{income$>$50K}  & sex                          & sex=Male                  \\
                                                  &                                    &                                & race                         & race=White                \\ \hline
            \multirow{2}{*}{COMPAS~\cite{compas}} & \multirow{2}{*}{7,214$\times$11}   & \multirow{2}{*}{no recidivism} & sex                          & sex=Female                \\
                                                  &                                    &                                & race                         & race=Caucasian            \\ \hline
            \multirow{2}{*}{German~\cite{german}} & \multirow{2}{*}{1,000$\times$21}   & \multirow{2}{*}{good credit}   & sex                          & sex=male                  \\
                                                  &                                    &                                & age                          & age$>$30                  \\ \hline
        \end{tabular}
    }
\end{table}

\subsection{Datasets \& Model}

\parh{Dataset.}~Our experiments are conducted on three real-world datasets:
Adult~\cite{adult}, COMPAS~\cite{compas}, and German~\cite{german}. These
datasets are widely used for fairness research~\cite{zhang2022adaptive,
chakraborty2018adversarial, chakraborty2020fairway, chakraborty2021bias,
li2022training}. \T~\ref{tab:data} shows the information of these datasets. Each
of them has two sensitive attributes, which smoothly enables analyzing the
trade-off between multiple sensitive attributes' fairness.

\parh{Model Training.}~Following Zhang et al.~\cite{zhang2022adaptive}, we use a
feed-forward neural network (FFNN) with five hidden layers. To adjust the
model's learning capacity, we control its size with a variable, \textit{model
width}. The default value of this variable is $4$, so layers of the model
contain $4\times 16=64$, $4\times8=32$, $4\times4=16$, $4\times2=8$, and
$4\times1=4$ neurons, respectively. For each dataset presented in
\T~\ref{tab:data}, we split the data into training and test sets with a ratio of
7:3. All trained models possess comparable performance~\cite{zhang2022adaptive,
chakraborty2021bias, peng2022fairmask, li2022training}. In particular, we
achieve 84.7\% accuracy on the Adult Income dataset, 67.4\% accuracy on the
COMPAS dataset, and 72.1\% accuracy on the German Credit dataset. We also
clarify that this model architecture is sufficient for our experiments, as all
three datasets contain a relatively small number of features (see
\T~\ref{tab:data}).

\begin{table}[!htbp]
    \centering
    \caption{Fairness improving methods.}
    \resizebox{\linewidth}{!}{
        \begin{tabular}{l|c}
            \hline
            \textbf{Category}                    & \textbf{Name}                                                       \\\hline
            \multirow{6}{*}{Pre-processing (6)}  & Reweighing~\cite{kamiran2012data}                                   \\\cline{2-2}
                                                 & Disparate Impact Remover (DIR)~\cite{feldman2015certifying}         \\\cline{2-2}
                                                 & FairWay~\cite{chakraborty2020fairway}                               \\\cline{2-2}
                                                 & FairSmote~\cite{chakraborty2021bias}                                \\\cline{2-2}
                                                 & FairMask~\cite{peng2022fairmask}                                    \\\cline{2-2}
                                                 & LTDD~\cite{li2022training}                                          \\\hline
            \multirow{3}{*}{In-processing (3)}   & Adversarial Debiasing (AD)~\cite{zhang2018mitigating}               \\\cline{2-2}
                                                 & Prejudice Remover (PR)~\cite{kamishima2012fairness}                 \\\cline{2-2}
                                                 & Exponentiated Gradient Reduction (EGR)~\cite{agarwal2018reductions} \\\hline
            \multirow{3}{*}{Post-processing (3)} & Reject Option Classification (ROC)~\cite{kamiran2012decision}       \\\cline{2-2}
                                                 & Equalized Odds (EO)~\cite{hardt2016equality}                        \\\cline{2-2}
                                                 & Calibrated Equalized Odds (CEO)~\cite{pleiss2017fairness}           \\\hline
        \end{tabular}
    }
    \label{tab:fair-methods}
\end{table}

\subsection{Fairness Improving Methods}

\T~\ref{tab:fair-methods} presents the fairness-improving methods used in
our experiments. It is notable that we have opted for significantly more
pre-processing methods than the other categories. This decision is
influenced by the trend in the software engineering community and the
greater compatibility offered by pre-processing methods. In particular, we
surveyed top-tier conferences/journals in the software engineering
community and found that pre-processing methods are generally dominant this
line of research. Moreover, these methods impose no constraints on the
model and have no impact on its output.

Conversely, some in-processing methods are designed for specific models, such as
FairNeuron~\cite{gao2022fairneuron} that only works on particular DNNs provided
by the authors. This makes in-processing less compatible and impedes a fair
comparison with other methods. Post-processing methods usually alter the model's
output, nullifying prediction probability and prohibiting robustness-related
analysis, such as adversary attacks. Given that said, as a comprehensive study,
we still select three representative methods from both in-processing and
post-processing methods.

\subsection{Metrics}
\label{subsec:metrics}

This paper employs a wide range of metrics to evaluate model performance,
fairness (both individual and group), and robustness. We present the
metrics used in our experiments in \T~\ref{tab:metrics} to ensure clarity.
For detailed definitions of each metric, interested readers may refer to
the cited references.

\begin{table}[!htbp]
    \centering
    \caption{Metrics information.}
    \resizebox{\linewidth}{!}{
        \begin{tabular}{l|c}
            \hline
            \textbf{Category}                        & \textbf{Name}                                                        \\\hline
            \multirow{2}{*}{Performance (2)}         & Accuracy (Acc)~\cite{aifclassificationmetric}                        \\\cline{2-2}
                                                     & F1 score (F1)~\cite{aifclassificationmetric}                         \\\hline
            \multirow{3}{*}{Group Fairness (3)}      & Disparate Impact (DI)~\cite{aifclassificationmetric}                 \\\cline{2-2}
                                                     & Statistical Parity Difference (SPD)~\cite{aifclassificationmetric}   \\\cline{2-2}
                                                     & Average Odds Difference (AOD)~\cite{aifclassificationmetric}         \\\hline
            \multirow{3}{*}{Individual Fairness (3)} & Consistency (Cons)~\cite{aifbinarylabeldatasetmetric}                \\\cline{2-2}
                                                     & Theil Index (TI)~\cite{aifbinarylabeldatasetmetric}                  \\\cline{2-2}
                                                     & Causal Discrimination Score (CDS)~\cite{zhang2022adaptive}           \\\hline
            \multirow{4}{*}{Robustness (4)}          & FGSM's Success Rate (FGSM)~\cite{goodfellow2014explaining}           \\\cline{2-2}
                                                     & PGD's Success Rate (PGD)~\cite{madry2017pgd}                         \\\cline{2-2}
                                                     & Rule-Based MI's Accuracy (Rule)~\cite{artmiattach} \\\cline{2-2}
                                                     & Black-Box MI's Accuracy (Bbox)~\cite{artmiattach}  \\\hline
        \end{tabular}
    }
    \label{tab:metrics}
\end{table}

Except for the Causal Discrimination Score (CDS), all fairness metrics are
computed by the widely-used AIF360~\cite{aif360-oct-2018}. We implement the
CDS metric by us. For robustness, we evaluate the model from two
perspectives: adversarial attack~\cite{goodfellow2014explaining,
madry2017pgd} and membership inference (MI) attack~\cite{artmiattach}. For the
adversarial attack, we use two standard approaches, the Fast Gradient Sign
Method (FGSM)~\cite{goodfellow2014explaining} and the Projected Gradient
(PGD)~\cite{madry2017pgd}. Here, the success rate is defined as the ratio
of the number of adversarial examples that successfully fool the model to
the total number of adversarial examples. In our experiments, the
FGSM/PGD implementation provided in Torchattacks~\cite{kim2020torchattacks}
is used. For the membership inference attack, we use two popular methods:
the rule-based and black-box methods. The rule-based method assumes that a
sample is a member if the model correctly predicts its label; otherwise,
the sample is a non-member. The black-box method trains a model to predict
whether a sample is a member or not. Both methods are implemented by
ART~\cite{art2018}.

Although users are typically only interested in metrics measured on the
test set, we also measure dataset properties (e.g., \textit{DI} of the
dataset) and metrics on the training set (e.g., \textit{SPD} model's
prediction on the training set). We regard these metrics as intermediate
nodes, such as when pre-processing methods alter the dataset's
properties, causing a change in the model's prediction on the training set,
which in turn causes a change in the model's prediction on the test set.
Therefore, these mediators contribute to causal graph learning. In
addition, we measure fairness metrics for two sensitive attributes,
respectively, to analyze the trade-off between multiple sensitive
attributes. Overall, the causal graph contains 46 nodes.

\section{Pilot Study on Causal Graph Quality}
\label{sec:pilot}

Before using the learned causal graph to answer RQs (\S~\ref{sec:evaluation}),
we study their accuracy through a pilot study.
This section consists of two pilot tasks: graph comparison and accuracy
verification. In the first task, we compare the six learned causal graphs (three
datasets $\times$ two sensitive attributes in each dataset) to find their
similarities and differences. In the second task, we conduct a human evaluation
and a quantitative analysis to verify the accuracy of the learned causal graphs.


\begin{figure}[!ht]
    \centering
    \includegraphics[width=0.99\linewidth]{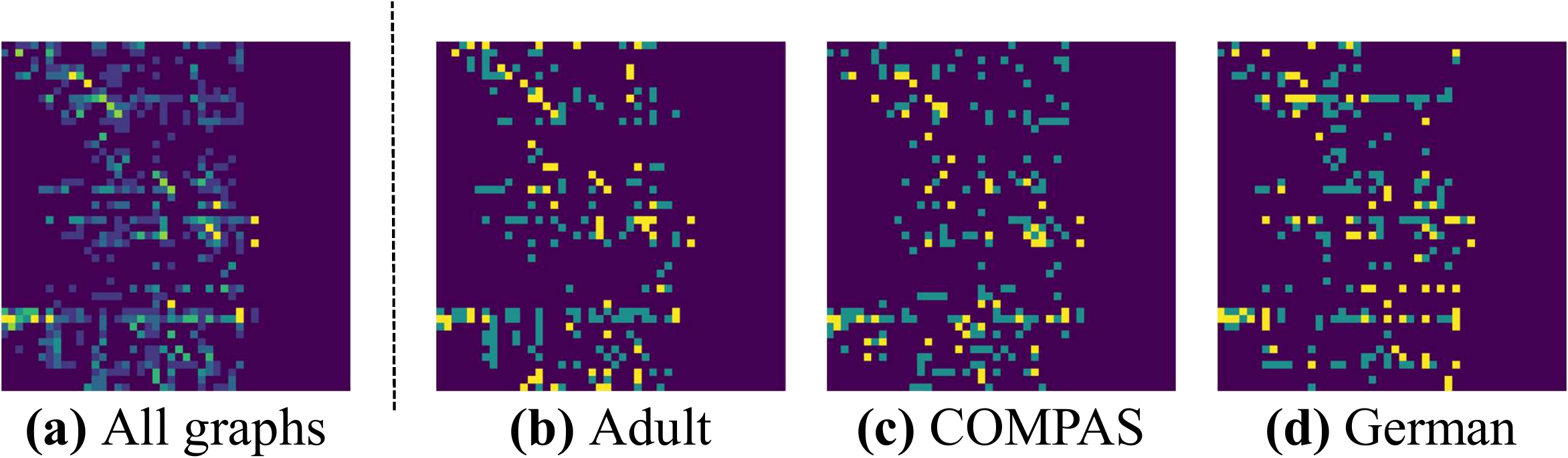}
    \caption{Overlap between the learned causal graph. The brighter the
        color, the higher the overlap.}
    \label{fig:overlap}
\end{figure}

\begin{figure}[!ht]
    \centering
    \includegraphics[width=0.50\linewidth]{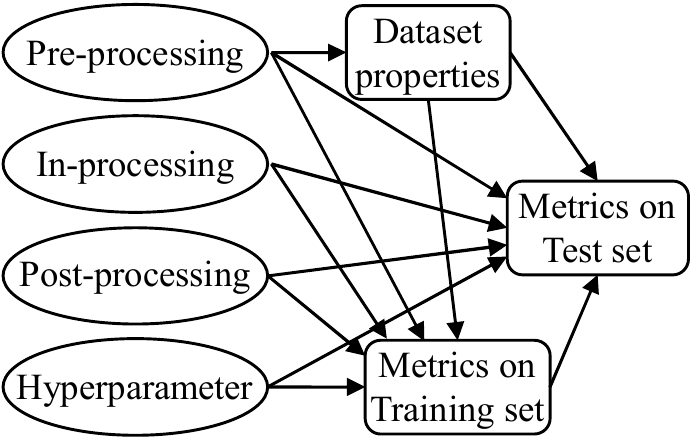}
    \caption{High-level common patterns across all graphs.}
    \label{fig:components}
\end{figure}

\subsection{Graph Comparison}



From \F~\ref{fig:overlap}(a), we can see that the overlap across all graphs
is small. These six graphs reach consensus on only 14 edges, while the
average number of edges per graph is 138.
According to previous research~\cite{baluta2022membership}, this enormous
distinction is a common phenomenon in causal discovery, compelling us to
learn six graphs instead of one in this study. We clarify that the
relationships among variables are typically more complex than we believe,
as there is no simple equation to characterize them (e.g., the relationship
between model loss and SPD score). Therefore, the causal graph may change
considerably if the dataset, model architecture, or even the sensitive
attribute addressed in this study is altered. 
In fact, as expected, when we retain the same dataset and model architecture, we
can observe that the overlap increases. In \F~\ref{fig:overlap}(b), (c), and
(d), the overlaps between two graphs range from 30\% to 50\%. 

Despite the intuitive observation, we clarify that the small overlaps between
the learned causal graphs, as reflected in \F~\ref{fig:overlap}, does not imply
that there is no common pattern across those graphs. For instance, we can
observe that the right parts of all graphs are ``empty'' (colored in dark
purple), signifying that nodes corresponding to those right parts lack parents.
This is because those nodes represent the fairness-improving methods. As
interventional variables, they are exogenous nodes in the graph. Based on this
kind of observation, we conclude a high-level common pattern across all graphs,
as shown in \F~\ref{fig:components}. This pattern is consistent with our
expectations, such that the pre-processing methods cause the change in
the data, which subsequently affect the model performance. Overall, we view
that the pilot study at this step demonstrates the high accuracy of the learned
causal graphs.

\subsection{Accuracy Verification}

As the study basis, the accuracy of the learned causal graphs is
essential to the effectiveness of the entire pipeline. Because of the absence of
ground truth, we propose a human evaluation to assess the quality and accuracy
of learned causal graphs. In particular, we invite six experts in the field of
software engineering (all of whom are Ph.D. students with extensive experience
in fairness or ML) to evaluate the causal graphs. 

Since there are 46 nodes in each graph, indicating high complexity and a large
number of edges, we presume that it is impractical to request that experts
construct causal graphs from scratch. Instead, the knowledge and experience of
experts are more suitable for validating the causal graphs learned by the causal
discovery algorithm.
Specifically, for each learned causal graph, we randomly sample three subgraphs
(each containing 15 nodes out of 46 nodes in total) and present them to the
experts for evaluation. Experts are requested to mark any edges on the subgraph
they disagree with and to note any edges not present in the subgraph that should
be included. Then, we gather this feedback and compute an \textit{error rate}
(i.e., the rate of incorrectly discovered edges by DiBS) and a
\textit{negative predictive value} (NPV; denoting the proportion of absent edges
in the learned graph). 

\begin{table}[!htbp]
    \centering
    \caption{Human evaluation.}
    \resizebox{0.9\linewidth}{!}{
        \begin{tabular}{l|l||cccc}
            \hline
            \multicolumn{2}{c}{}    & nodes & edges & error rate & NPV              \\ \hline
            \multirow{2}{*}{Adult}  & sex   & 46    & 141        & 1.78\%  & 6.84\% \\
                                    & race  & 46    & 136        & 4.77\%  & 4.15\% \\ \hline
            \multirow{2}{*}{COMPAS} & sex   & 46    & 130        & 10.19\% & 2.30\% \\
                                    & race  & 46    & 133        & 7.90\%  & 3.39\% \\ \hline
            \multirow{2}{*}{German} & sex   & 46    & 141        & 6.71\%  & 3.69\% \\
                                    & age   & 46    & 146        & 5.73\%  & 2.58\% \\ \hline
        \end{tabular}
    }
    \label{tab:human}
\end{table}

The results are shown in \T~\ref{tab:human}. We can see that the error rate is
less than 10\% in the vast majority of instances (with only one exception that
trivially exceeds 10\% by 0.19\%), where the lowest value is only 1.78\%. The
results for NPV are even more impressive. In all cases, the NPV is less than
7\%, with a minimum value of 2.30\%. This result indicates that the learned
causal graphs are highly accurate and can serve as the foundation for further
analysis.

\begin{table}[!htbp]
    \centering
    \caption{Ablation study results.}
    \resizebox{\linewidth}{!}{
        \begin{tabular}{l|l||ccccc}
            \hline
            \multicolumn{2}{c}{}    & Full ver. & w/o pre & w/o in & w/o post & w/o all        \\ \hline
            \multirow{2}{*}{Adult}  & sex       & 1.00    & 0.56   & 0.64     & 0.77    & 0.00 \\
                                    & race      & 1.00    & 0.43   & 0.81     & 0.82    & 0.00 \\ \hline
            \multirow{2}{*}{COMPAS} & sex       & 1.00    & 0.55   & 0.90     & 0.89    & 0.00 \\
                                    & race      & 1.00    & 0.48   & 0.89     & 0.88    & 0.00 \\ \hline
            \multirow{2}{*}{German} & sex       & 1.00    & 0.65   & 0.54     & 0.76    & 0.00 \\
                                    & age       & 1.00    & 0.60   & 0.67     & 0.81    & 0.00 \\ \hline
        \end{tabular}
    }
    \vspace{-5pt}
    \label{tab:ablation}
\end{table}

In addition to the human evaluation, we also report the statistics of the
learned causal graphs and their ablated versions in \T~\ref{tab:ablation}, where
``Full ver.'' represents that all fairness-improving methods are included, and
``w/o XX'' indicates that some of the fairness-improving methods are excluded
(e.g., ``w/o pre'' means pre-processing methods are excluded). We use normalized
Bayesian Gaussian equivalent (BGe) score~\cite{geiger1994learning,
geiger2002parameter}, whose implementation is provided by
DiBS~\cite{lorch2021dibs}, to measure the quality of the learned causal graph.
As a standard metric in causal discovery, BGe scores reflect the data-fitness of
the graph. The higher the BGe score, the better the learned graph. Note that BGe
scores are data-specific, i.e., they cannot be compared across different
scenarios, so we only compare the BGe scores on the same row in
\T~\ref{tab:ablation}.

The results indicate that the application of fairness-improving techniques
has a substantial impact on the quality of the learned causal graphs.
Furthermore, each category of fairness-improving methods is essential to
the quality of the learned causal graph, as removing any category of
fairness-improving methods will result in a significant drop in the BGe scores.

\section{Evaluation}
\label{sec:evaluation}

We now investigate the mechanisms underlying various trade-off phenomena
related to fairness, including the trade-off between fairness and model
performance (\textbf{RQ1}), the trade-off between multiple sensitive
attributes (\textbf{RQ2}), and the trade-off between fairness and model
robustness (\textbf{RQ3}). In each RQ, we first present the discovered
trade-off (with counts of occurrences) and corresponding causes that are
revealed by \A~\ref{alg:analysis} using the learned graphs. Then, we
analyze the causes and discuss the implications of the trade-off and their
causes.

For the sake of space, metric names are abbreviated as follows:
\textit{prefix}-\textit{metric}, where \textit{prefix} is either
\textit{Tr} (measured on training data), \textit{Te} (measured on testing
data), or \textit{D} (datasets' properties), and \textit{metric} follows
the same naming convention as in \T~\ref{tab:metrics}. Additionally,
\textit{Width} denotes the width of the neural network, which is the
hyperparameter that controls the number of neurons in each layer. Using
\texttt{Tr-SPD} as an illustration, it represents ``SPD'' (Statistical
Parity Difference; see \T~\ref{tab:metrics}) measured on the training set.
Note that the term ``causes'' refers exclusively to the metrics mentioned
in \S~\ref{subsec:metrics}. All fairness-improving methods are excluded from
``causes'', as they are undoubtedly viewed as the causes of the trade-offs
triggered by themselves. We clarify that the purpose of this study is to
analyze the mechanisms underlying the trade-offs, so we concentrate on the
intricate relationships among metrics.


\parh{Processing Time.}~Building each causal graph requires
approximately 30 minutes to learn from the data. As for the trade-off analyses
(depicted in \A~\ref{alg:analysis}), we report that the average processing time
for analyzing each trade-off is less than 1 minute, the majority of which is
spent on computing the ATEs. In studying the following RQs, the processing time
of computing ATEs is negligible, typically less than 20 seconds.

\begin{figure*}[!htbp]
    \centering
    \includegraphics[width=0.85\linewidth]{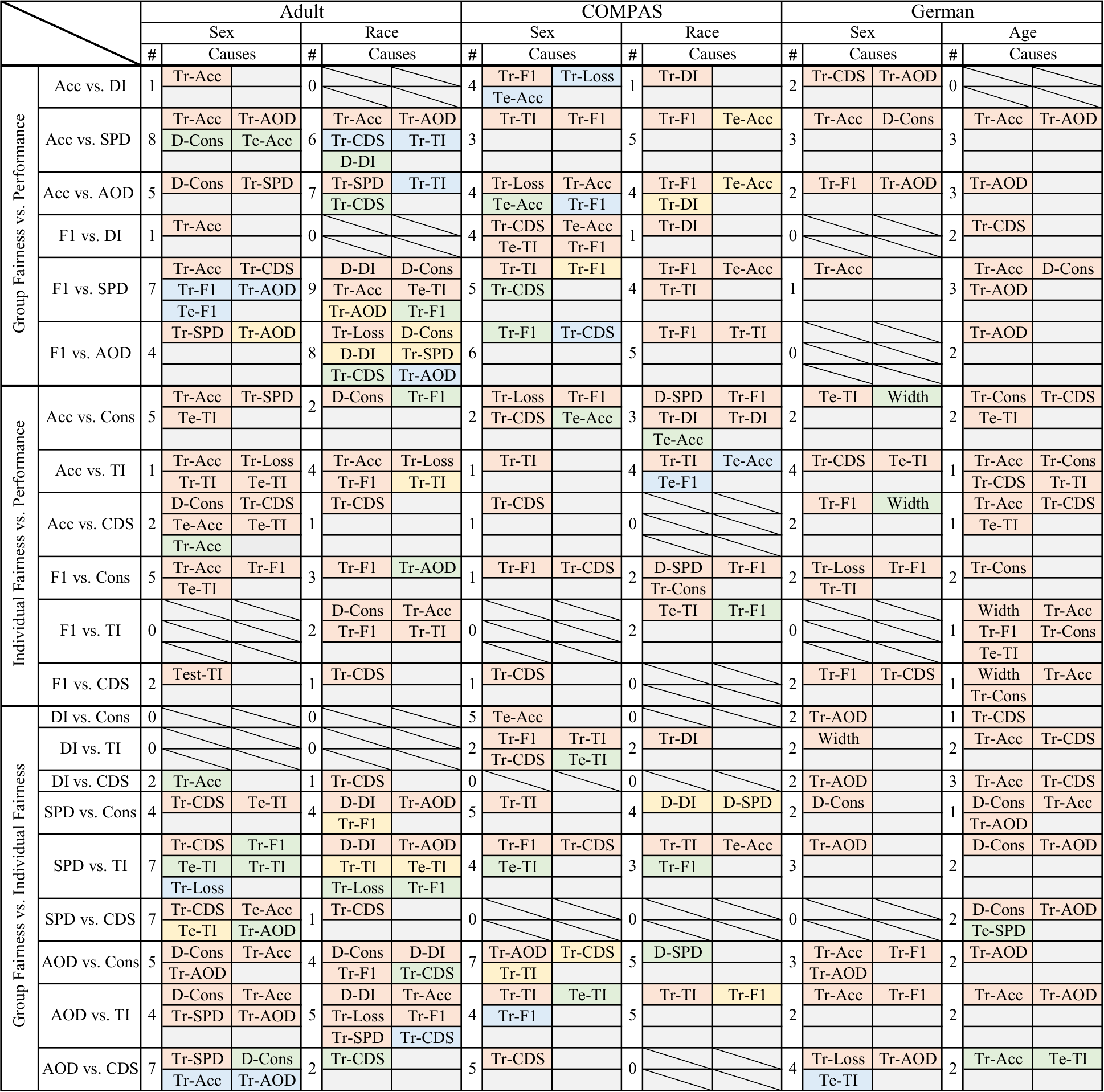}
    \caption{Trade-offs between fairness and model performance. The color
        of each cell indicates the confidence of revealed causes. In
        particular, \colorbox{redMark}{red} indicates a full confidence
        (100\%), \colorbox{yellowMark}{yellow} indicates a high confidence
        (70\%--99\%), \colorbox{greenMark}{green} indicates a medium
        confidence (30\%--69\%), and \colorbox{blueMark}{blue} indicates a
        low confidence (1\%--29\%).}
    \label{fig:rq1}
\end{figure*}

\subsection{RQ1: Fairness vs. Model Performance}

As mentioned in \S~\ref{sec:background}, there are two types of fairness: group
fairness and individual fairness. This section explores the trade-offs between
both types of fairness and model performance, i.e., group fairness vs. model
performance and individual fairness vs. model performance. Additionally, we also
investigate the trade-offs between group fairness and individual fairness, as
this is a topic that has been primarily focused in the fairness
literature~\cite{speicher2018unified, hort2021fairea, zhang2022adaptive}.

\F~\ref{fig:rq1} reports the observed trade-offs for each scenario in the form
of ``counts (\#)'' and ``causes''. The ``counts'' column shows the trigger time
of trade-offs w.r.t. fairness-improving methods in \T~\ref{tab:fair-methods}.
For instance, the ``counts'' of ``Acc vs. DI'' is $1$, when ``sex'' is the
addressed sensitive attribute on the ``Adult'' dataset. This means that there is
only one fairness-improving method that triggers the trade-off between Accuracy
and DI on Adult. The ``causes'' column shows the causes of the trade-offs, which
are revealed by \A~\ref{alg:analysis}. For each cause, we also report the
``confidence'' (i.e., the percentage of the trade-off caused by the metric) in
four colors. 
Additionally, we use a diagonal line when no causes are found.


\F~\ref{fig:rq1} provides a comprehensive view of the trade-offs between
fairness and model performance by listing all observed trade-offs. As the count
of the majority of trade-offs is not zero, this table demonstrates that
``trade-off'' is a common phenomenon in fairness-improving methods. Moreover,
the kinds of trade-offs vary considerably across different scenarios. For
example, in the case of group fairness vs. performance, we observe that the
trade-offs observed on Adult are mainly between SPD and performance metrics (the
trade-off between Accuracy and SPD is observed eight times, and the trade-off
between F1 and SPD is observed seven times when sex is the addressed sensitive
attribute). In contrast, on COMPAS, most trade-offs are observed between AOD and
performance metrics. This result suggests that the selection of fairness metrics
may have a substantial impact on the number and type of trade-offs that can be
observed.

We presume that the more frequently a metric is observed as a cause, the more
important the metric is, as it is more likely to reveal how well the
fairness-improving method works with regard to the trade-off, i.e., whether this
method achieves a win-win situation. Based on this presumption, we expect to
identify the most informative and beneficial metric for the development of fair
ML based on the results of our experiments. From \F~\ref{fig:rq1}, we observe
that there is no unified cause for all trade-offs. Instead, the causes of
trade-offs vary considerably across different scenarios. Furthermore, the
distribution of causes is not uniform. In \F~\ref{fig:rq1}, the most prevalent
causes of trade-offs are metrics measured on the training set (e.g.,
\texttt{Tr-TI}). This type of metric functions as a cause for trade-offs 190
times, which is far more than other types of causes. In comparison, metrics
measured on the test set (e.g., \texttt{Te-TI}) are only 38 times the cause of
trade-offs, and for metrics of datasets' properties, the number is 29. This
significant difference motivates us to investigate the distribution of causes
further.

\begin{table}[!htbp]
    \centering
    \caption{Distribution of causes in RQ1.}
    \resizebox{0.80\linewidth}{!}{
        \begin{tabular}{ll|lll|l}
            \hline
            \multicolumn{2}{c}{}        & Data & Train & Test & Total      \\\hline
            \multirow{3}{*}{Group}      & DI   & 8     & 6    & 0     & 14 \\\cline{2-6}
                                        & SPD  & 4     & 8    & 1     & 13 \\\cline{2-6}
                                        & AOD  & N/A   & 30   & 0     & 30 \\\hline
            \multirow{3}{*}{Individual} & Cons & 17    & 6    & 0     & 23 \\\cline{2-6}
                                        & TI   & N/A   & 21   & 21    & 42 \\\cline{2-6}
                                        & CDS  & N/A   & 36   & 0     & 36 \\\hline
            \multicolumn{2}{l|}{Total}  & 29   & 107   & 22   & 158        \\\hline
        \end{tabular}
    }
    \label{tab:distribution}
\end{table}

\T~\ref{tab:distribution} presents the distribution of causes. In this table, we
report the frequency of each fairness metric being the causes of trade-offs
listed on \F~\ref{fig:rq1}.\footnote{Here, we only report fairness metrics,
because the target of this experiment is to identify the most informative and
beneficial fairness metrics for the development of fair ML.} The column
indicates the phase in which the metric is measured, i.e., ``Data'' for the
dataset's properties, ``Train'' for the training set, and ``Test'' for the test
set. Note that AOD, TI, and CDS depend on the model's prediction, so they are
N/A for the ``Data'' column. This table reveals that AOD is the most common
cause of trade-offs among all group fairness metrics, occurring 30 times. For
individual fairness, TI is the most common cause of trade-offs, observed 42
times. Also, individual fairness metrics generally cause trade-offs more
frequently than group fairness metrics.

\parh{RQ1 Findings:}~In this RQ, we have the following two suggestions for users
and developers of fair ML.

\begin{enumerate}[leftmargin=*,noitemsep,topsep=0pt]
    \item For the users, they should be aware that the results of
          experiments for fairness-improving methods may be biased by the
          selection of metrics. Moreover, the metrics that work well in one
          scenario to discover the trade-offs may not work well in other
          scenarios. To obtain a more comprehensive and faithful
          understanding of the fairness-improving method, we recommend that
          users use sufficient metrics or our causality analysis-based
          method, to explore the presumably optimal choice of metrics for
          their scenarios.

    \item For the developers, we suggest that they pay closer attention
          to the metrics measured on the training set, as they are the most common
          causes of trade-offs according to our study. Additionally, we recommend AOD
          among all group fairness metrics and TI among all individual fairness
          metrics.
\end{enumerate}

\subsection{RQ2: Multiple Sensitive Attributes}

In this RQ, we investigate the trade-offs between multiple sensitive
attributes. For the sake of space, we only report the results of
experiments on COMPAS, and present other results on the
website~\cite{website}. To distinguish metrics measured on different
sensitive attributes, we change the abbreviation of metrics to
``\textit{prefix}-\textit{sensitive attribute}-\textit{metric}''. For
example, ``\texttt{Tr-Sex-SPD}'' represents ``SPD'' measured on the
training set with ``Sex'' as the sensitive attribute. In this RQ, only the
trade-offs between group fairness for different sensitive attributes are
considered, as individual fairness metrics are not related to the sensitive
attribute.

\begin{figure}[!htbp]
    \centering
    \includegraphics[width=1.0\linewidth]{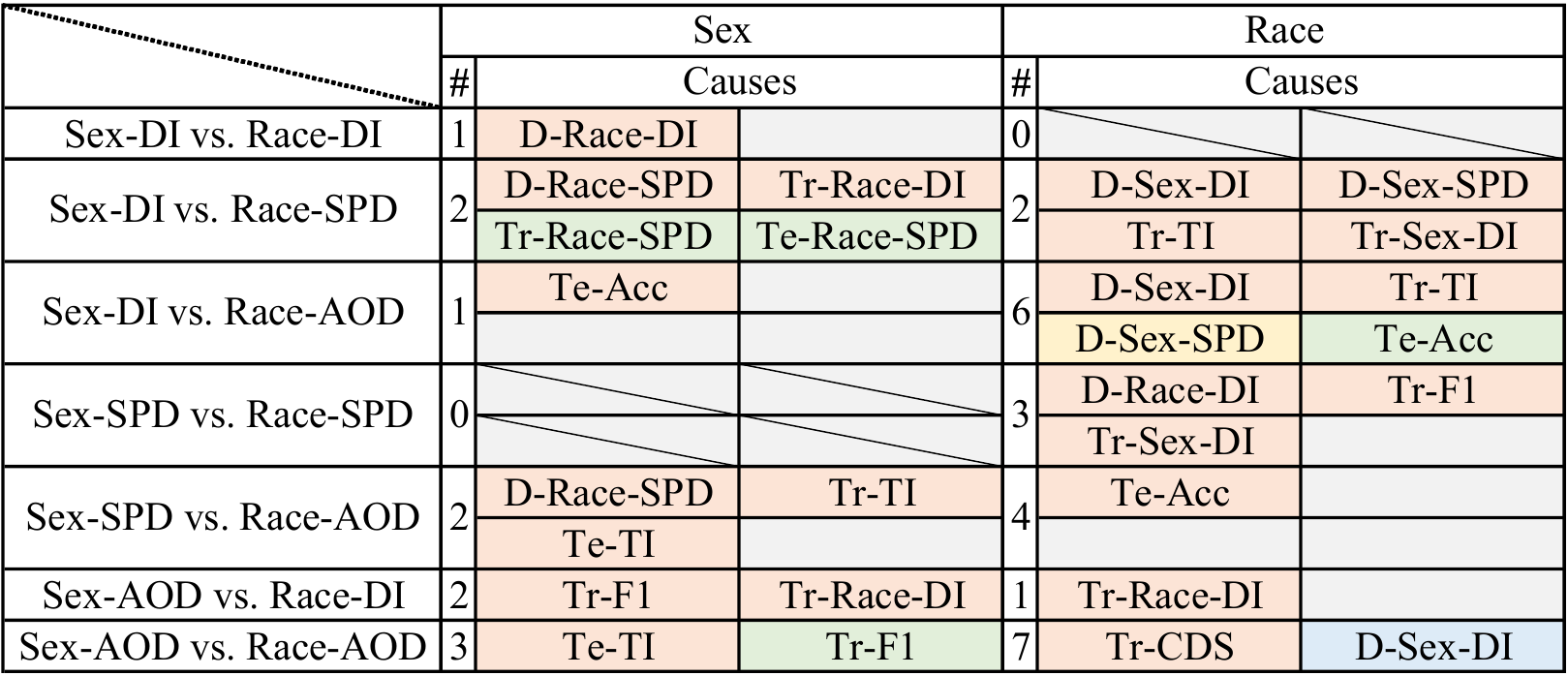}
    \caption{Trade-offs between multiple sensitive attributes.}
    \label{fig:rq2}
\end{figure}

\begin{figure}[!htbp]
    \centering
    \includegraphics[width=0.95\linewidth]{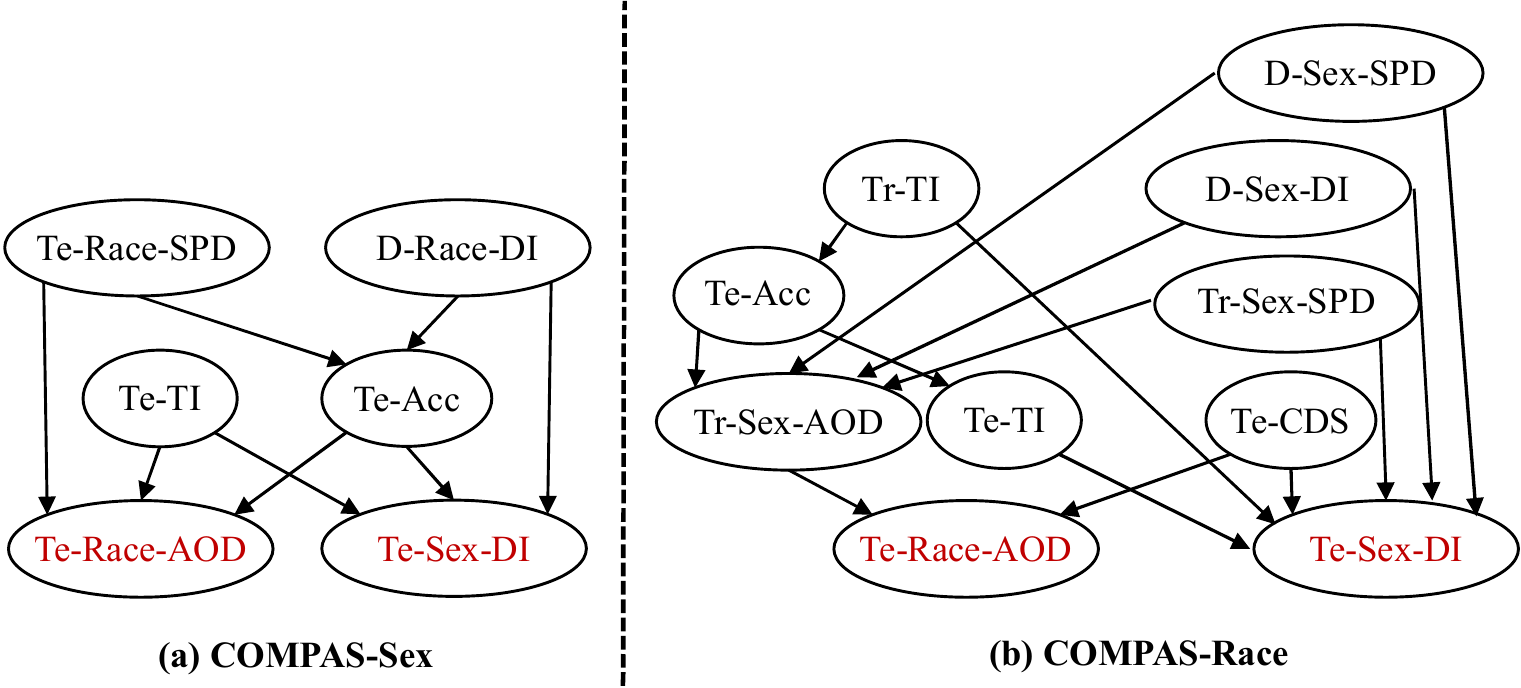}
    \caption{A comparison between trade-offs in different scenarios.}
    \label{fig:rq2-eg}
\end{figure}

\F~\ref{fig:rq2} presents the results of experiments on COMPAS. The left
part of this table shows the results when the addressed sensitive attribute
of the fairness-improving method is set to ``Sex'', and the right part
shows the results when the sensitive attribute is set to ``Race''. Although
these two parts attain consensus in some cases (e.g., they both agree that
the trade-off between ``\texttt{Sex-DI}'' and ``\texttt{Race-DI}'' is
rare), they have several substantial differences. For example, the left
part reveals that the trade-off is rare between ``\texttt{Sex-DI}'' and
``\texttt{Race-AOD}'' (the ``counts'' is only one), whereas the right part
shows that the trade-off is frequent between these two metrics (six
counts). To explain this difference, we present the causal graphs of these
two metrics in \F~\ref{fig:rq2-eg}.

This figure provides an intuitive explanation for the aforementioned
distinction. Clearly, the causal graph of the scenario ``COMPAS-Race'' is
more intricate than that for ``COMPAS-Sex''. With more common ancestors in
the graph, it is more likely that more causes leading to trade-offs will be
identified. In addition, \F~\ref{fig:rq2-eg} also explain the difference in
found causes between ``COMPAS-Sex'' and ``COMPAS-Race''. In
\F~\ref{fig:rq2-eg}(a), the metric ``\texttt{Te-Acc}'' not only has direct
causal relations with ``\texttt{Sex-DI}'' and ``\texttt{Race-AOD}'', but
also mediates the effect from ``\texttt{Te-Race-SPD}'' and
``\texttt{D-Race-DI}'' to ``\texttt{Sex-DI}'' and ``\texttt{Race-AOD}''.
Therefore, ``\texttt{Te-Acc}'' is the cause with full confidence here. In
contrast, ``\texttt{Te-Acc}'' no longer has direct causal relation with
either ``\texttt{Sex-DI}'' or ``\texttt{Race-AOD}'' in
\F~\ref{fig:rq2-eg}(b). This explains why its confidence decreases
drastically in scenario ``Compas-Race''.

\parh{RQ2 Findings:}~We observe the substantial variation in patterns of
trade-offs between different sensitive attributes even on the same dataset.
Furthermore, we take a pair of causal graphs as an example to explain the
difference in detail.

\begin{figure}[!htbp]
    \centering
    \includegraphics[width=0.95\linewidth]{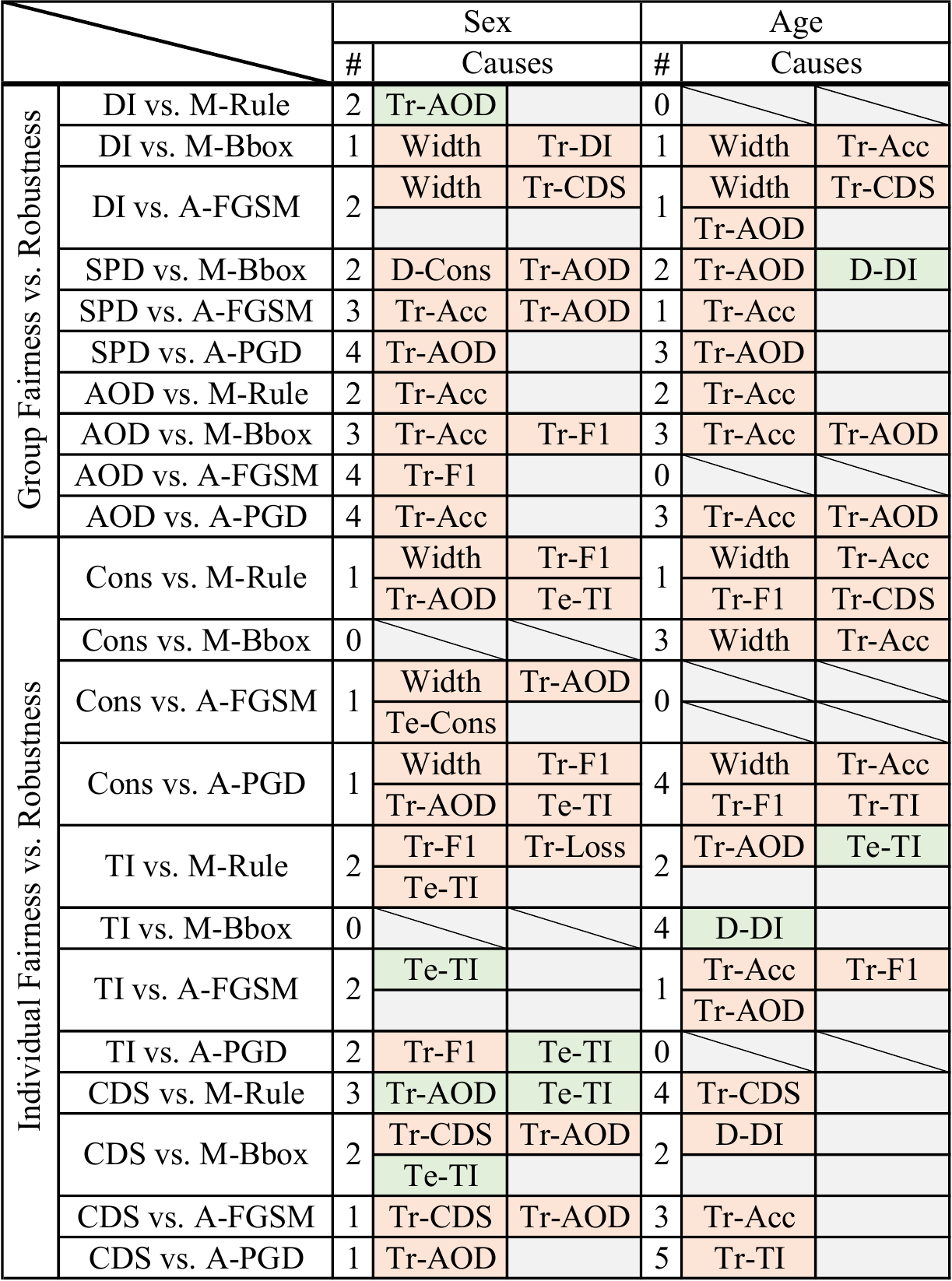}
    \caption{Trade-offs between fairness and model robustness.}
    \label{fig:rq3}
\end{figure}
\subsection{RQ3: Fairness vs. Model Robustness}

As essential properties of ML models, both fairness and robustness receive
considerable attention in the research community. Although some
works~\cite{xu2021robust, sun2022towards, chai2023robust} examine them
simultaneously, no one has systematically studied the trade-offs between
them. This RQ investigates the trade-offs between fairness and model
robustness. Similar to RQ2, we only report the results of experiments
conducted on German in \F~\ref{fig:rq3} and present the remaining results
on the website~\cite{website}. For the abbreviation of robustness metrics,
we use ``\textit{prefix}-\textit{robustness metric}'' to represent them,
where \textit{prefix} is either ``A'' (adversarial attack) or ``M''
(membership inference).

Comparing patterns of trade-offs in \F~\ref{fig:rq3} with those in
\F~\ref{fig:rq1}, an apparent distinction is that the hyperparameter
``Width'' causes much more trade-offs than it does in \F~\ref{fig:rq1}'s
same scenario (``German-Sex'' and ``German-Age''). We interpret this
difference as reasonable because the robustness metrics are more sensitive
to the model's learning ability and the degree of overfitting. Hence, it is
not unexpected to see that performance metrics, including Accuracy and F1
score, also cause more trade-offs in \F~\ref{fig:rq3}. This result
is consistent with research on membership
inference~\cite{shokri2017membership, salem2018ml, baluta2022membership}.

It is evident to conclude that trade-offs between fairness and robustness are
inevitable, given that fairness-improving techniques typically have a
significant impact on the model's performance. Moreover, the large number of
observed trade-offs in \F~\ref{fig:rq3} further suggests that the trade-offs
between fairness and robustness should be taken seriously. Therefore, we suggest
that future research on fairness-improving methods should consider this kind of
trade-off, and faithfully disclose the results of potential robustness
downgrades. We also clarify that the additional effort required to consider
robustness will not be excessive, as \F~\ref{fig:rq3} shows that the recommended
metrics in RQ1 (AOD among group fairness metrics and TI among individual
fairness) are still highly effective to inspect trade-offs between fairness and
robustness.

\parh{RQ3 Findings:}~Fairness and robustness contain inevitable trade-offs.
We advocate taking into account robustness metrics when designing
fairness-improving methods, and we have illustrated that the extra cost is
moderate.

\section{Downstream Application}
\label{sec:application}


As detailed in \S~\ref{sec:evaluation}, our method delivers a comprehensive and
in-depth understanding of the trade-offs among multiple metrics. Naturally, the
identified causal graph provides valuable insights into the selection of the
optimal fairness-improving method for a given scenario. As a ``by-product'', in
this section, we present a case study that highlights the versatile application
of our method in fairness-improving approach selection. 

The case study is conducted on all datasets mentioned in \T~\ref{tab:data},
specifically examining the interplay of accuracy, SPD, and consistency as
key factors to consider in selecting a suitable fairness-improving method.
Specifically, we first identify the causes of the trade-offs between the
selected metrics. Then, we find the optimal value of each cause using ATE,
which constitutes the optimal setting for fairness-improving methods. We
compare our method against the Adaptive Fairness Improvement
(AFI)~\cite{zhang2022adaptive}, a state-of-the-art approach, which is
tailored for this task and is not designed for trade-off analysis.

\begin{table}[!htbp]
    \centering
    \caption{Comparison of AFI vs.~our method on fairness-improving method
        selection. Each cell of AFI and ours reports the difference compared to
        the default model (w/o FI). \textcolor{myGreen}{Green} and
        \textcolor{myBrown}{brown} indicate an
        \textcolor{myGreen}{improvement}/\textcolor{myBrown}{downgrade},
        respectively.} \resizebox{1.0\linewidth}{!}{
        \begin{tabular}{|l|l||c|c|c|c|c|c|}
            \hline
                                       &        & \multicolumn{2}{c|}{Adult} & \multicolumn{2}{c|}{COMPAS} & \multicolumn{2}{c|}{German}                                                                                  \\\cline{3-8}
                                       &        & Sex                        & Race                        & Sex                         & Race                     & Sex                      & Age                      \\\hline
            \multirow{3}{*}{Acc}       & w/o FI & .8500                      & .8489                       & .6727                       & .6740                    & .7233                    & .7197                    \\
                                       & AFI    & \textcolor{myBrown}{-.0101}    & \textcolor{myBrown}{-.0167}     & \textcolor{myBrown}{-.0032}     & \textcolor{myBrown}{-.0050}  & \textcolor{myBrown}{-.0046}  & \textcolor{myBrown}{-.0050}  \\
                                       & Ours   & \textcolor{myBrown}{-.0138}    & \textcolor{myBrown}{-.0205}     & \textcolor{myGreen}{+.0098}    & \textcolor{myGreen}{+.0072} & \textcolor{myGreen}{+.0100} & \textcolor{myGreen}{+.0310} \\\hline
            \multirow{3}{*}{$|$SPD$|$} & w/o FI & .1730                      & .0974                       & .1575                       & .1609                    & .0685                    & .0915                    \\
                                       & AFI    & \textcolor{myGreen}{-.0608}   & \textcolor{myGreen}{-.0299}    & \textcolor{myGreen}{-.1161}    & \textcolor{myGreen}{-.0926} & \textcolor{myGreen}{-.0228} & \textcolor{myGreen}{-.0321} \\
                                       & Ours   & \textcolor{myGreen}{-.1718}   & \textcolor{myGreen}{-.0936}    & \textcolor{myGreen}{-.1488}    & \textcolor{myGreen}{-.1397} & \textcolor{myGreen}{-.0658} & \textcolor{myGreen}{-.0548} \\\hline
            \multirow{3}{*}{Cons}      & w/o FI & .9600                      & .9593                       & .9080                       & .9098                    & .7914                    & .8036                    \\
                                       & AFI    & \textcolor{myBrown}{-.0236}    & \textcolor{myBrown}{-.0101}     & \textcolor{myBrown}{-.0074}     & \textcolor{myBrown}{-.0061}  & \textcolor{myBrown}{-.0013}  & \textcolor{myBrown}{-.0100}  \\
                                       & Ours   & \textcolor{myGreen}{+.0179}   & \textcolor{myGreen}{+.0050}    & \textcolor{myGreen}{+.0081}    & \textcolor{myGreen}{+.0151} & \textcolor{myGreen}{+.0215} & \textcolor{myGreen}{+.0364} \\\hline
        \end{tabular}
    }
    \label{tab:discussion}
\end{table}



We report the evaluation results in \T~\ref{tab:discussion}. In particular, we
find that our method surpasses AFI in almost all scenarios. This is reasonable:
AFI can only select a single fairness-improving method, while our method
can effectively combine multiple fairness-improving methods using the causal
graph. For example, for the \texttt{German-Age} scenario, we found that the
optimal combination of fairness-improving methods was to use both disparate
impact remover (DIR) and predictive rate (PR), with their respective ratios set
to 0.6 and 0.2. However, the limitation of AFI results in reduced effectiveness.

\section{Related Work}
\label{sec:related}


\parh{Trade-off Study in ML Fairness.}~Prior works have investigated the
trade-offs associated with fairness. The analyzed trade-offs include
fairness vs. accuracy~\cite{corbett2017algorithmic, speicher2018unified,
hort2021fairea, dutta2020there, kim2020fact, zhang2021ignorance,
tizpaz2022fairness}, group fairness vs. individual
fairness~\cite{friedler2016possibility, binns2020apparent}, and fairness
vs. robustness~\cite{xu2021robust, sun2022towards, chai2023robust}.
Typically, these studies have three primary goals: (1) establishing the
theoretical existence of trade-offs, (2) designing methods to achieve
optimal trade-offs, and (3) identifying the best trade-off through
empirical comparisons. No one has, however, systematically analyzed the
influence of fairness-improving methods over ML pipelines, as measured by
the metrics in \S~\ref{subsec:metrics}, to cast light on the causes of
trade-offs.


\parh{Causality Analysis in SE.}~Recent years have witnessed a growing
interest in applying causality analysis to SE. The high interpretability of
causal graphs makes them appealing for a variety of SE problems, including
software configuration~\cite{dubslaff2022causality}, root cause
analysis~\cite{johnson2020causal, kuccuk2021improving, he2022perfsig}, and
deep learning testing/repairing~\cite{sun2022causality, zhang2022adaptive,
ji2023cc}. We have compared with AFI in \S~\ref{sec:application} to
highlight the distinct focus and superior performance.



\section{Threat to Validity}


\parh{Internal Validity.}~The increase in the number of nodes
in the causal graph may facilitate the discovery of causal relations, but
it has a substantial impact on the computational cost. To balance the
trade-off, we select 12 widely-used fairness-improving methods and 12
representative metrics. We presume that the number of nodes is sufficient
to guarantee the accuracy of our analysis. Regarding users of the proposed
analysis framework, they are encouraged to tailor the selection of nodes to
their specific requirements. In terms of causal discovery, we employ DiBS,
a state-of-the-art causal discovery algorithm, to infer causal graphs.
Although DiBS outperforms other algorithms~\cite{lorch2021dibs}, it may not
always identify the true causal graphs. To mitigate this threat, we use
human evaluation in our pilot study for validating the derived causal
graphs. In the future, we will attempt to combine graphs learned by one or
multiple state-of-the-art algorithms to further improve the quality of
causal graphs.

\parh{External Validity.}~Our study focuses on neural networks, possibly
limiting generalizability to other ML models. However, given the popularity
of neural networks and their strong compatibility with numerous
fairness-improving methods, we argue that our results hold considerable
value. To further alleviate this threat, we conduct experiments across
various network architectures and datasets.

\section{Conclusion}

This research analyzes the trade-offs among multiple factors in fair ML via
causality analysis. We propose a set of design principles and optimizations to
facilitate an effective usage of causality analysis in this field. With
extensive empirical analysis, we establish a comprehensive understanding of the
interactions among fairness, performance, and robustness. 

\section*{Acknowledgement}

We thank anonymous reviewers for their valuable feedback. We also thank the
participants in the human evaluation for their time and effort. We
acknowledge Timothy Menzies for his insightful suggestions regarding the
quality of learned causal graphs. The HKUST authors are supported in part
by the HKUST-VPRDO 30 for 30 Research Initiative Scheme under the contract
Z1283. Yanhui Li is supported by the National Natural Science Foundation of
China (Grant No. 62172202).

\bibliographystyle{IEEEtran}
\bibliography{bib/causality,bib/machine-learning,bib/ref,bib/testing-cv,bib/fair}

\end{document}